\def\year{2020}\relax
\documentclass[letterpaper]{article} 
\usepackage{aaai20}  
\usepackage{times}  
\usepackage{helvet} 
\usepackage{courier}  
\usepackage[hyphens]{url}  
\usepackage{graphicx} 
\usepackage{float}
\usepackage{enumitem}
\usepackage{amsmath}
\usepackage{amssymb}
\usepackage{multirow, makecell}
\usepackage{color}

\DeclareMathOperator*{\softmax}{softmax}

\usepackage{graphicx}  
\title{LATTE: Latent Type Modeling for Biomedical Entity Linking}

\author{Ming Zhu,\textsuperscript{\rm 1}\thanks{Work done while MZ was an intern at AWS AI.} 
Busra Celikkaya,\textsuperscript{\rm 2} 
Parminder Bhatia,\textsuperscript{\rm 2} 
Chandan K. Reddy\textsuperscript{\rm 1}\\
\textsuperscript{\rm 1}Department of Computer Science, Virginia Tech, Arlington, VA 22203\\ 
\textsuperscript{\rm 2}AWS AI, Seattle, WA 98121\\ 
mingzhu@vt.edu, \{busrac, parmib\}@amazon.com, reddy@cs.vt.edu 
}

\begin{document}

\maketitle

\begin{abstract}

Entity linking is the task of linking mentions of named entities in natural language text, to entities in a curated knowledge-base. This is of significant importance in the biomedical domain, where it could be used to semantically annotate a large volume of clinical records and biomedical literature, to standardized concepts described in an ontology such as Unified Medical Language System (UMLS). We observe that with precise type information, entity disambiguation becomes a straightforward task. However, fine-grained type information is usually not available in biomedical domain. Thus, we propose \textbf{LATTE}, a \textbf{LAT}ent \textbf{T}ype \textbf{E}ntity Linking model, that improves entity linking by modeling the latent fine-grained type information about mentions and entities. Unlike previous methods that perform entity linking directly between the mentions and the entities, LATTE jointly does entity disambiguation, and latent fine-grained type learning, without direct supervision. We evaluate our model on two biomedical datasets: MedMentions, a large scale public dataset annotated with UMLS concepts, and a de-identified corpus of dictated doctor's notes that has been annotated with ICD concepts. Extensive experimental evaluation shows our model achieves significant performance improvements over several state-of-the-art techniques.
\end{abstract}

\section{Introduction}
\label{sec:introduction}

\noindent With the advancements in the healthcare domain, we have witnessed a considerable increase in the amount of biomedical text, including electronic health records, biomedical literature and clinical trial reports \cite{reddy2015healthcare}. To successfully utilize the wealth of knowledge contained in these records, it is critical to have automated semantic indexing techniques. \textit{Entity linking} 
refers to the process of automatically linking mentions of entities in raw text, to a standardized list of entities in a knowledge-base. This process typically requires two steps. First, all the mentions of entities in the raw text are annotated using a standard Named Entity Recognition (NER) technique \cite{lample2016neural}. Next, the extracted mentions are linked to the corresponding entities in the entity disambiguation stage. Although a significant amount of work has been done in the domain of entity linking for text found on the web, where the objective is to link the mentions to standard knowledge-bases such as Freebase \cite{bollacker2008freebase}, most of the techniques cannot be directly transferred to biomedical domain, which poses a number of new challenges to the entity linking problem.

\begin{figure}[h]
    \centering
    \includegraphics[ width=.95\columnwidth]{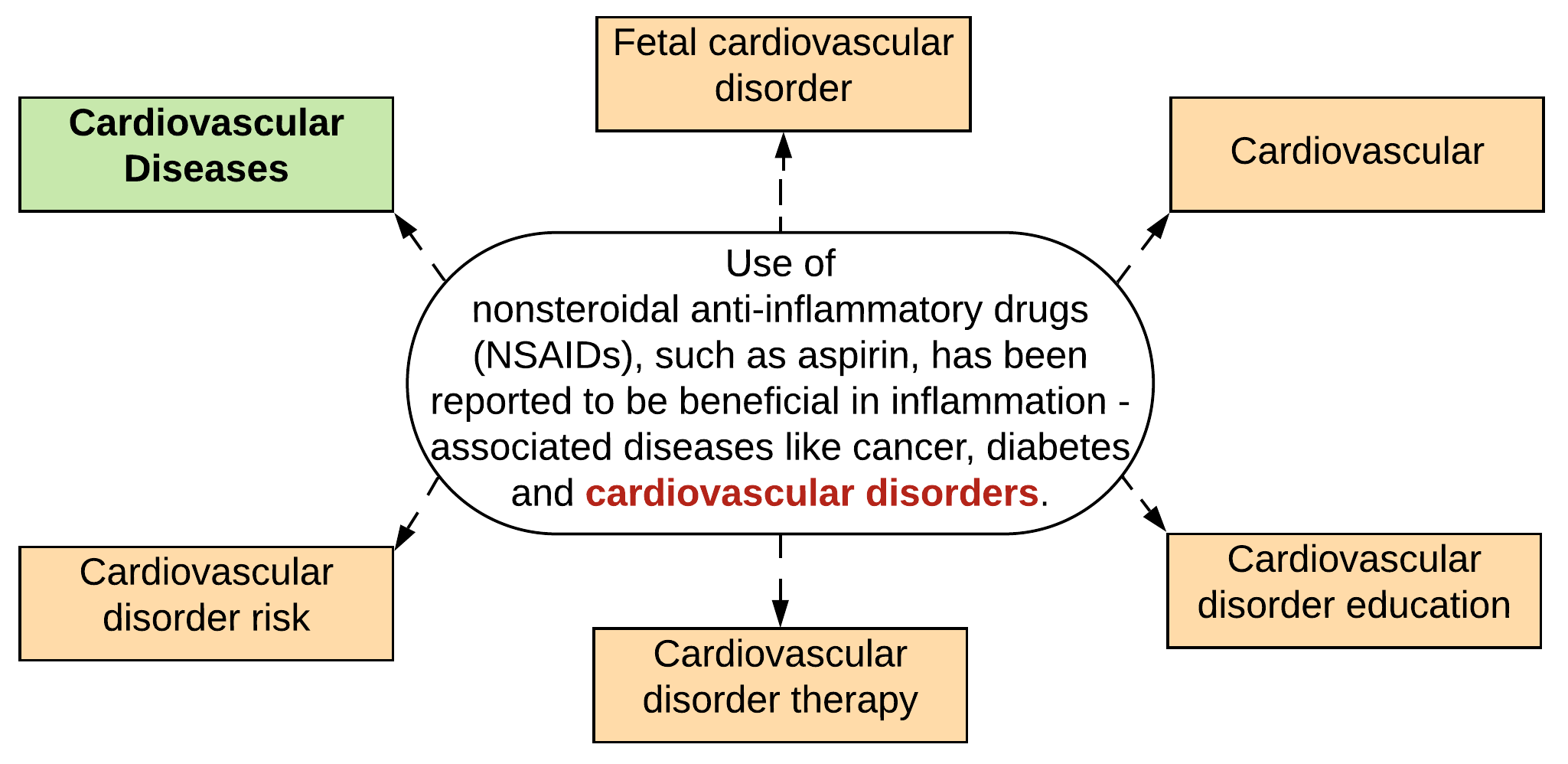}
    \caption{An example of biomedical entity linking. Phrase shown in red is the extracted mention, the orange boxes refer to the top candidate entities retrieved from the biomedical knowledge-base, and the green box is the ground truth entity for this mention. }
    \label{fig:el_example}
\end{figure}

Biomedical entity linking is the task of linking mentions in biomedical text, such as clinical notes, or biomedical literature, to medical entities in a standard ontology such as Unified Medical Language System (UMLS) \cite{bodenreider2004unified}. In the healthcare domain, accurate entity disambiguation is crucial to the understanding of biomedical context. Many distinct biomedical concepts can have very similar mentions, and failure in disambiguation will lead to incorrect interpretation of the entire context. This will introduce huge risks in medical-related decision making. Moreover, biomedical entity linking can be useful in many other applications, which require automatic indexing of the text. For instance, it can be used by healthcare providers to automatically link the medical records of patients to different medical entities, which can then be used for downstream tasks such as diagnosis/medication decision making, population and health analytics \cite{bhatia2019comprehend}, predictive modeling \cite{jin2018improving}, medical information retrieval, information extraction \cite{hoffmann2011knowledge}, and question answering \cite{yih2015semantic}. 

Entity linking on biomedical text differs from that on other general domains of text, such as web documents, in many ways. Consider the example in Figure \ref{fig:el_example}, where \textit{cardiovascular disorders} is a mention of the entity \textit{Cardiovascular Diseases}, and others are the top candidate entities retrieved from UMLS. 
\begin{itemize}[leftmargin=*]
    \item First, the mentions can be ambiguous. In this example, almost all the other candidates have words that exactly match those in the mention. If we only use surface level features, it will be hard to link the mention to the correct entity. This requires the model to have a good semantic understanding of the mention and its context. 
    \item Second, the candidates can be similar with each other, not only in surface, but also in semantic meaning. In many cases it requires additional information, such as fine-grained types, to distinguish the correct entity. 
    \item Another challenge in medical entity linking is that the mentions and the context are usually longer in length, as compared to general domain. This makes the traditional entity linking techniques less effective on medical text. 
    \item Finally, medical text contains many domain specific terminologies as well as abbreviations and typos. Thus, many terms cannot be found in standard pre-trained embeddings such as GloVe \cite{pennington2014glove}, and it makes neural models less effective due to a large number of out-of-vocabulary words.
\end{itemize}

\begin{figure}[h]
    \centering
    \includegraphics[ width=.95\columnwidth]{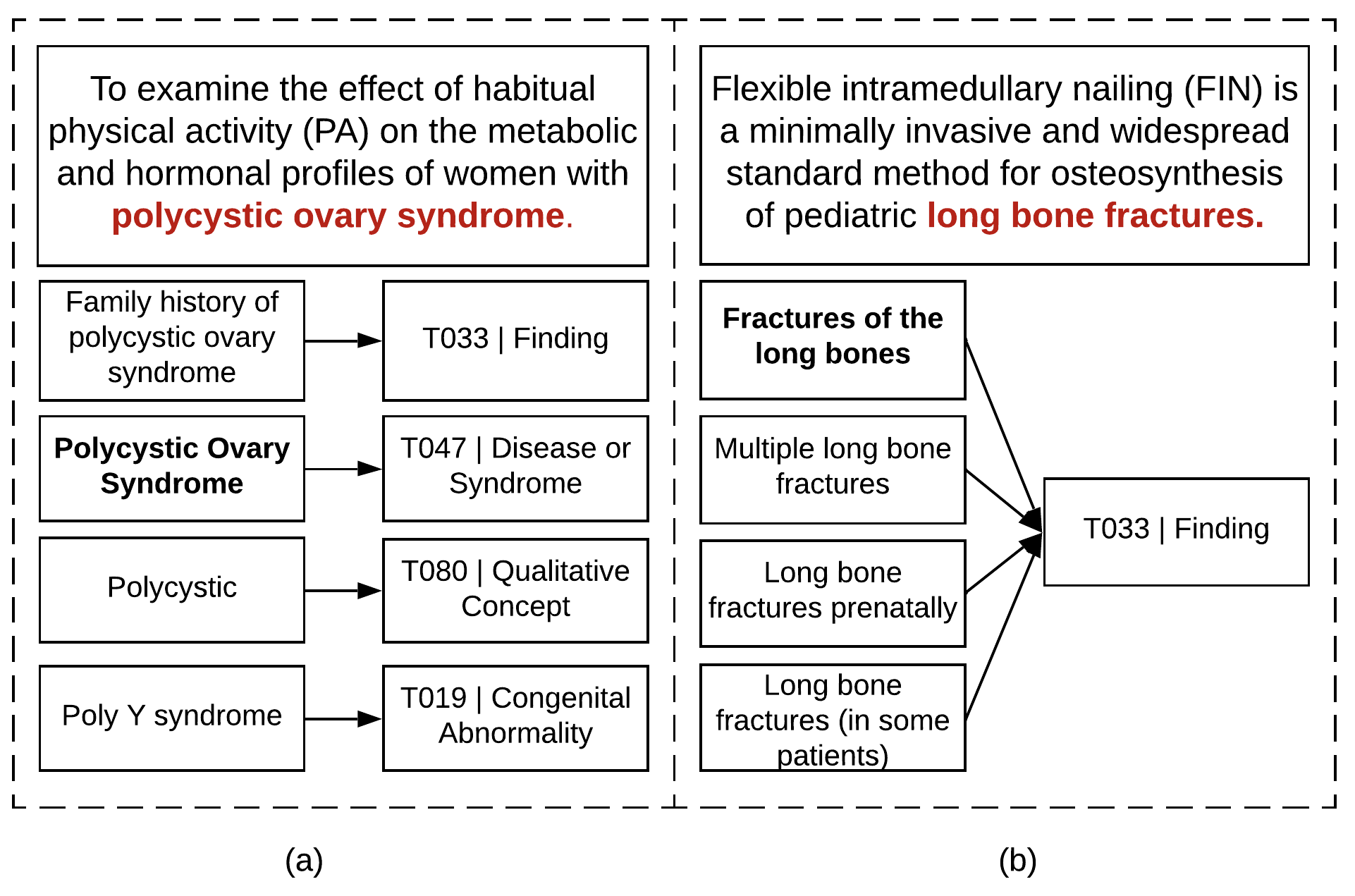}
    \caption{Examples of biomedical entity linking with type information.}
    \label{fig:fine_grained_entity_type}
\end{figure}

A key observation in the process of entity linking is that if we have the fine-grained types of mentions in the raw text, and types of entities in the knowledge-base, entity disambiguation becomes much easier. For example, in Figure \ref{fig:fine_grained_entity_type}\textit{(a)}, each candidate entity has different semantic type from UMLS Semantic Network \cite{mccray1989umls}. If we can infer the correct mention type, which in this case most likely is \textit{Disease or Syndrome}, we can make the correct linking decision with no further effort. However, the type information in biomedical domain is not always available, and the available ones are usually far from fine-grained. 

Taking into account all these challenges, in this work, we propose \textbf{LATTE} (\textbf{Lat}ent \textbf{T}ype \textbf{E}ntity Linking model), a novel neural network based model for entity linking in the biomedical text. LATTE introduces the concept of latent-type modeling for entities and their mentions. The \textit{latent} types refer to the implicit attributes of each entity. To guide the training process, we also use the coarse-grained \textit{known} entity types as auxiliary supervision. 
To further enable our model to link the mentions with the entities from the knowledge-base, we use an attention-based mechanism, that equips the model to rank different candidate entities for a given mention, based on their semantic representations. 
We evaluate the performance of our model using a large scale entity linking dataset from the biomedical domain and a de-identified corpus of doctor's notes, against several state-of-the-art baselines.  

The rest of this paper is organized as follows: Section \ref{sec:related_work} provides an overview of some of the existing techniques related to our model. In Section \ref{sec:model}, we describe our proposed model along with the details about the optimization and training process. In Section \ref{sec:experimental_results}, we give the details about our experimental results including the evaluation metrics and baseline models. Finally, Section \ref{sec:conclusion} concludes the paper with possible directions for future work.

\section{Related Work}
\label{sec:related_work}

\subsection{Neural Entity Linking}
Neural entity linking has attracted significant interest from researchers for many years. Most of the existing techniques in this domain can broadly be classified into three categories. \textit{Context modeling} approaches model the context of mentions and the candidate entities at different levels of granularity to get the similarity between them. An example of such approaches is \cite{francis2016capturing}, which uses a set of vectors that include mention, mention context and the document that mention appears in, and vectors from entity article title and document, to compute the mention and entity representations, respectively.  \cite{gupta2017entity} also extensively makes use of context information in the entity linking process. \textit{Type modeling} approaches make use of the entity types in the linking process. This is based on the observation that if the entity types are known, entity linking performance can be improved significantly \cite{raiman2018deeptype}. \textit{Relation modeling} approach models the latent relations between different mentions without direct supervision \cite{le2018improving}. These relations can be used as a measure of coherency of the linking decisions, and can ultimately guide the entity linking process.

\subsection{Neural Network Models for Text Matching}
With the success of deep learning, many neural network based models have been proposed for semantic matching, and document ranking. Models such as ARC-I \cite{hu2014convolutional} first compute the representation of the two sentences, and then compute their relevance. The representations computed in this manner cannot incorporate the pairwise word-level interactions between the sentences. Interaction based models, such as Deep Relevance Matching Model (DRMM) \cite{guo2016deep} and  MatchPyramid \cite{pang2016text}, compute the interactions between the words from two sequences, and then compute the relevance based on these features. The models proposed in \cite{xiong2017end} and \cite{dai2018convolutional} use kernel pooling on interaction features to compute similarity scores, followed by convolutional layers to compute the relevance score. The HAR model \cite{zhu2019hierarchical} uses a hierarchical attention mechanism to rank answers for healthcare-related queries.

\section{The Proposed Model}
\label{sec:model}

\subsection{Motivation} 
We already showed in section \ref{sec:introduction}, that with precise entity type information (Figure \ref{fig:fine_grained_entity_type}\textit{(a)}), entity disambiguation becomes a straightforward task. However, such detailed information is not usually available. For example, in the UMLS Semantic Network \cite{mccray1989umls}, there are only 127 types in total, while UMLS has about 900,000 unique entities \cite{bodenreider2004unified}. In general, most known types are far from fine-grained. Furthermore, manually labeling all the entities for precise types requires significant amount of resources and can be a daunting task. 
Therefore, we are motivated to model the latent fine-grained types for all the entities in the knowledge-base without direct supervision. \\

\noindent \textbf{Latent Fine-grained Types:} We argue that fine-grained types do exist. For example, the entities \textit{Type 2 Diabetes Mellitus} and \textit{Parkinson Disease} both have the semantic type \textit{Disease or Syndrome}, but the former is a metabolic disorder, while the latter is a nervous system disorder. In this case, the finer-grained type can be the body system where the disease occur. Similarly, in Figure \ref{fig:fine_grained_entity_type}\textit{(b)}, for mention \textit{long bone fractures}, all the candidates share the same semantic type \textit{Finding}, yet they still have different intrinsic attributes which can be used to distinguish them from others. Here we see the intrinsic attributes as finer-grained types for each entity. Moreover, since there is no fixed set of fine-grained types, we do not have ground truth labels for them. This motivates us to model the fine-grained types as latent variables, and we model them using different constraints. \\

\noindent \textbf{Binary Pairwise Relation Constraint:} One constraint is the binary pairwise relation between mention and each candidate entity. Specifically, if one candidate is the ground truth entity for a given mention, the relation between them is labeled as $1$; otherwise $0$. We can learn the latent types from this pairwise information, as a mention and its ground truth candidate should share the same latent type distribution. Alternatively, we can see the pairwise relation label as a similarity measure between the mention and the candidates, and we can infer how similar the latent types of a mention and a candidate are from this similarity measure. \\

\noindent \textbf{Type Hierarchy Constraint: } Additionally, we can make use of the coarse-grained known types. Note that (1) known types can be of any kind and not necessarily semantic types from the UMLS Semantic Networks; (2) regardless of the number of known types, we consider them as coarse-grained, as we can always model finer-grained types. The known types are usually generic in nature, and they can be further divided into sub-types. We can view these sub-types as the previously mentioned latent fine-grained types. Thus we introduce a hierarchy in the types: the known types are the top-level nodes in the hierarchy, and the latent fine-grained types are the low-level nodes. Therefore, we can supervise on the known types to model the latent types. \\

\noindent \textbf{Multi-tasking of Entity Linking and Type Classification: } To model the latent types with both the Pairwise Constraint and the Type Hierarchy, we simultaneously optimize for both entity linking and type classification in our model. The entity linking module uses attention mechanism to obtain a similarity score between a mention-candidate pair, and is supervised on the pairwise relation labels. The type classification module consists of two type classifiers: one for mention, and the other for candidates. Both classifiers are supervised on the known type labels, and the weights are shared between them. The similarity of the two output latent type distributions is used as another mention-candidate similarity score. This score is combined with the previous score to obtain the final similarity score. By jointly optimizing the two tasks, we expect the entity linking performance to improve. 

\subsection{Problem Statement} Given a mention phrase (mention with context) $p$ from a text in the biomedical domain, and a set of candidate entities $C=$ \{$c_1, .., c_l$ \} from a knowledge-base, the model computes a relevance score $r_{p,c}$ for each entity in $C$, based on its relevance with the mention.

\begin{figure*}[!ht]
\centering
\includegraphics[width=.95\textwidth]{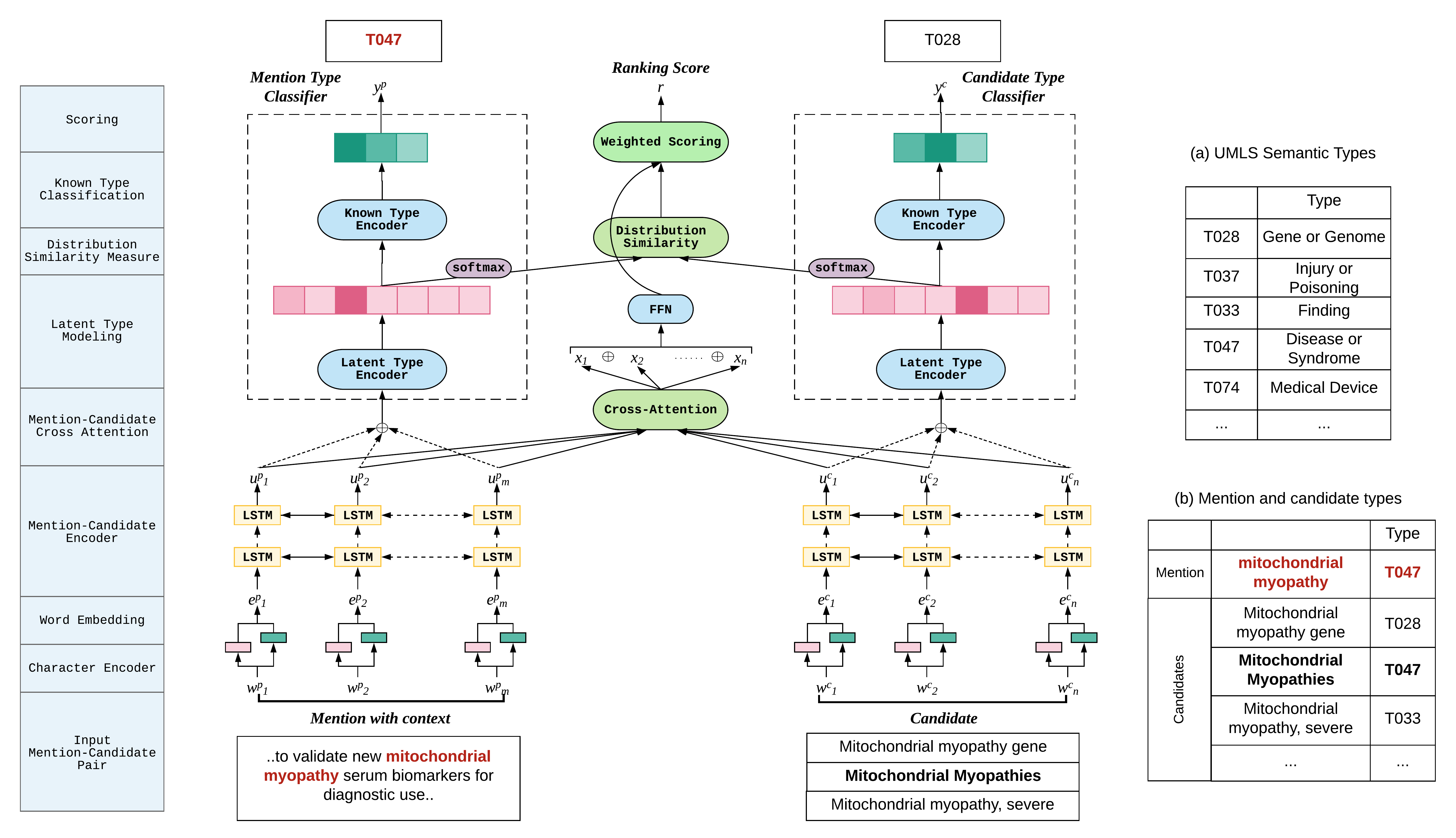}
\caption{The overall architecture of the proposed LATTE model for biomedical entity linking. Table (a) shows part of the UMLS Semantic Types, which we use as the known types. Table (b) shows the type information of the mention and candidates in the given example. }
\label{fig:model_architecture}
\end{figure*}

\subsection{Model Architecture}
Various components of our model are described in detail below. The overall architecture of the model is illustrated in Figure \ref{fig:model_architecture}. For all notations, we use the superscripts $p$ and $c$ for mention and candidate sequences respectively, where $m$ and $n$ denote their corresponding sequence lengths.
\\ \\
\noindent \textbf{Embedding Layer:} The first layer in our model is the embedding layer. This layer takes as input the word tokens $\{{w^p_i}\}_{i=1}^m$ and $\{{w^c_i}\}_{i=1}^n$ for the mention and the candidate sequences respectively, and returns the embedding vectors $\{{e^p_i}\}_{i=1}^m$ and $\{{e^c_i}\}_{i=1}^n$ for each word token. To overcome the problem of out-of-vocabulary words, we use a combination of word and character embeddings. First, the character embeddings for each character in a word are concatenated and passed through a convolutional neural network. The resultant vector is then concatenated with the word embedding, obtained from pre-trained embeddings like GloVe \cite{pennington2014glove} to get the word representation.
\\ \\
\noindent \textbf{Encoder:} To get a contextual representation of the words, we use a multi-layer Bidirectional LSTM \cite{hochreiter1997long} encoder for both the mention and the candidate sequences. This layer takes the word representations from the embedding layer as the input, and returns the contextual representations $\{{u^p_i}\}_{i=1}^m$ and $\{{u^c_i}\}_{i=1}^n$ of words in the two sequences. The resultant vectors have the contextual information from both the backward and the forward context encoded in them.
\begin{equation}
    \begin{split}
    \overrightarrow{u^p_i} = \overrightarrow{LSTM}(\overrightarrow{u^p}_{i-1}, e^p_i) \ \ & \ \
    \overleftarrow{u^p_i} = \overleftarrow{LSTM}(\overleftarrow{u^p}_{i+1}, e^p_i) \\
    \overrightarrow{u^c_i} = \overrightarrow{LSTM}(\overrightarrow{u^c}_{i-1}, e^c_i) \ \ & \ \
    \overleftarrow{u^c_i} = \overleftarrow{LSTM}(\overleftarrow{u^c}_{i+1}, e^c_i) \\
    u^p_i = [\overrightarrow{u^p_i}; \overleftarrow{u^p_i}] \ \ & \ \ 
    u^c_i = [\overrightarrow{u^c_i}; \overleftarrow{u^c_i}]
    \end{split}
\end{equation}

\noindent \textbf{Cross-Attention Layer:} This layer computes the interaction between the mention and the candidate vectors. It takes their encoded representations as the input, and computes the relevance between each pair of the mention and candidate word vectors, generated by the encoder layer. We use a bidirectional attention mechanism, as proposed in \cite{seo2016bidirectional}, for this layer. Specifically, we first compute the similarity matrix $S \in \mathbb{R}^{m \times n}$ between $\{{u^p_i}\}_{i=1}^m$ and $\{{u^c_i}\}_{i=1}^n$. Each element $s_{ij}$ of this matrix is calculated as follows:
\begin{equation}
    \begin{split}
    s_{ij} = w^{T}_{a} \cdot & [u^c_i; u^p_j; u^c_i \odot u^p_j]
    \end{split}
\end{equation}

\noindent After this, we compute the mention-to-candidate attention $S^{\alpha}$, and candidate-to-mention attention $S^{\beta}$ as
\begin{equation}
    \begin{split}
    S^{\alpha} = \softmax_{row} & (S)\text{,}   \\
    \bar{S}^{\beta} = \softmax_{col} (S)\text{, and}\ \ \ \
    & S^{\beta} = S^{\alpha} \cdot \bar{S}^{\beta^T}\text{.} \\
    \end{split}
\end{equation}

\noindent Finally, attended vectors $\{x_j\}_{j=1}^n$ can be computed as
\begin{equation}
    \begin{split}
    a^{\alpha}_j = \sum_{i} s^{\alpha}_{ij} u^c_i \text{,}\ \ \ \ & \ \ \ \ 
    a^{\beta} _j = \sum_{i} s^{\beta}_{ij} u^p_i \text{,}\\
    x_j = [u^p_j; a^{\alpha}_j; & u^p_j \odot a^{\alpha}_j; u^c_j \odot a^{\beta} _j]\text{.}
    \end{split}
\end{equation}

\noindent All the attended vectors from the cross-attention layer are then concatenated to form $X = [x_1; .., x_n]$, and fed into a multi-layer feed-forward network, to obtain the attention-based relevance score $f$ between the two sequences,
\begin{equation}
    f = ReLU (w_f \cdot X + b_f)\text{.}
\end{equation}

\noindent \textbf{Latent Type Similarity:} This layer takes the output states of the encoder layer, and then concatenates them to form a fixed-dimensional vector $u^p = [u^p_1; .. ;u^p_m]$ for the mention, and $u^c = [u^c_1; .. ;u^c_n]$ for the candidate. The two vectors are then passed through feed-forward layers, followed by a softmax layer to obtain two probability distributions over $k$ latent types. 
It then computes the similarity $g$ between the two distributions of the mention and the candidate using a standard distance metric like cosine similarity:
\begin{equation}
    \begin{split}
    v^p = w_{l} \cdot u^p + b_{l}\text{,} & \ \ \ \
    \hat{v^p} = \softmax(v^p)\text{,} \\
    v^c = w_{l} \cdot u^c + b_{l}\text{,} & \ \ \ \
    \hat{v^c} = \softmax(v^c)\text{,} \\
     g = & \frac{\hat{v^p} \cdot \hat{v^c}}{||\hat{v^p}|| \ ||\hat{v^c}||}\text{.}
    \end{split}
\end{equation}

\noindent \textbf{Known Type Classifier:} To incorporate the known type information and to indirectly supervise the latent type modeling, we introduce known type classifier, which is trained to predict the entity types of both the mention and candidate vectors. It takes the encoded representations $v^p$ and $v^c$ of the latent types, and then uses a feed-forward network with Rectifier Linear Unit (ReLU) activation, to predict their known types $y^p$ and $y^c$, respectively. 
\begin{equation}
    \begin{split}
    & y^p = ReLU (w_{k} \cdot v^p + b_{k}) \\
    & y^c = ReLU (w_{k} \cdot v^c + b_{k})
    \end{split}
\end{equation}

\noindent \textbf{Ranking Layer:} After computing the interaction score $f$, and the latent type similarity $g$, we use the ranking layer to obtain the relevance score between the mention and the candidate sequences. This module performs a weighted-combination of the two relevance scores, to compute the final relevance score $r$.
\begin{equation}
    r = w_r^f \cdot f + w_r^g \cdot g
\end{equation}
Here, $w_r^f$ and $w_r^g$ are learnable weights.

\subsection{Optimization}

\noindent Our model incorporates two objectives, one for the type prediction, and another for candidate scoring. We jointly optimize these two objectives during our training process. \\

\noindent \textbf{Type Classification loss:} To incorporate the knowledge about the \textit{known} categorical types into the semantic representation of mentions and the entities, we minimize the categorical cross-entropy loss. Given the known type $y \in \{y^p, y^c\}$ of a mention or a candidate, and its predicted type distribution $\hat{y}$, the loss is calculated as follows:

\begin{equation}
    \mathcal{L}^{type} = - \sum_{j=1}^K y_j \log (\hat{y_j})
\end{equation}

\noindent \textbf{Mention-Candidate Ranking loss:} For a given mention, we want to ensure that the correct candidate $c_{pos}$ gets a higher score compared to the incorrect candidates $c_{neg}$. Hence, we use max-margin loss as the objective function for this task. Given the final scores $r_{p, c_{pos}}$ and $r_{p, c_{neg}}$ of $p$ with respect to $c_{pos}$ and $c_{neg}$ respectively, the ranking loss is calculated as follows:

\begin{equation}
    \mathcal{L}^{rank} = \max\{0, M- r_{p, c_{pos}} + r_{p, c_{neg}}\}
\end{equation}

\section{Experimental Results}
\label{sec:experimental_results}

\subsection{Datasets}
We use two datasets to evaluate the performance of the proposed model. MedMentions  \cite{mohan2019medmentions} contains 4392 abstracts from PubMed, with biomedical entities annotated with UMLS concepts. It also contains up to 127 semantic types for each entity from the UMLS Semantic Network \cite{mccray1989umls}, which we use for the known type classification. We also use a de-identified corpus of dictated doctor's notes, which we refer to as 3DNotes. It is annotated with problem entities related to signs, symptoms and diseases. These entities are mapped to the 10th version of International Statistical Classification of Diseases and related health problems (ICD-10), which is part of UMLS. The annotation guidelines are similar to the i2b2 challenge guidelines for the problem entity \cite{uzuner20112010}. We use the top categories in the ICD-10 hierarchy as the known types. 
For both datasets, we take 5 words before and after a mention as the mention context.

\begin{table}[ht]
\centering
\begin{tabular}{p{1.2cm} p{2.1cm} p{1cm} p{1cm} p{1cm} }
\Xhline{3\arrayrulewidth}
\textbf{Dataset} & \textbf{Statistics} & \textbf{Train} & \textbf{Dev} & \textbf{Test} \\
\hline

\multirow{3}{1.2cm}{Med Mentions} &
\#Documents & 2,635 & 878 & 879 \\
& \#Mentions & 210,891 & 71,013 & 70,364 \\
& \#Entities & 25,640 & 12,586 & 12,402 \\
\hline
\multirow{3}{1.2cm}{3DNotes} &
 \#Documents & 2,133 & 525 & 745 \\
& \#Mentions & 22,266 & 5,373 & 8,065 \\
& \#Entities & 2,026 & 1,030 & 1,209 \\
\hline
\Xhline{3\arrayrulewidth}
\end{tabular}
\caption{Statistics of the datasets used. Note that the "\#Entities" refers to the number of unique entities.}
\label{tab:stats_medment}
\end{table}

\subsection{Candidate Generation} For MedMentions, we follow the approach of candidate generation described in \cite{murty2018hierarchical}. 
We take only the top 9 most similar entities (excluding the ground truth entity) as the negative candidates. In addition, the ground truth entity will be considered as the positive candidate, thus forming a set of 10 candidates for each mention. For 3DNotes, we use a similar approach to generate candidates from ICD-10.

\subsection{Evaluation Metrics}
To evaluate the proposed model, we measure its performance against the baseline techniques using Precision@1 (the precision when only one entity is retrieved) and Mean Average Precision (MAP). These metrics were chosen considering the fact that our problem setup is a ranking problem. 
Note that, in our case, since each mention has only one correct candidate entity, Precision@1 is also equivalent to Recall@1.

\subsection{Implementation Details}
We implemented our model and all other baselines in PyTorch \cite{paszke2017automatic}. The model was trained using Adam optimizer \cite{kingma2014adam}, with a learning rate of $10^{-4}$. We used GloVe embeddings with 300 dimensions as the input word vectors, and the output dimension of the character CNN was 512, making each word a 812-dimensional vector. The encoders used two Bi-LSTM layers, where the output dimension of each individual LSTM layer was 512. The number of latent types, $k$, is set to 2048. The hyperparameter values were obtained based on the experimental results on the validation set.

\subsection{Baselines}
For the quantitative evaluation of the proposed LATTE model, we use the following state-of-the-art baseline methods for comparison.

\begin{itemize}[leftmargin=*]
    \item \textbf{TF-IDF:} This is a standard baseline for NLP tasks. Here, we use character level $n$-grams as the terms, with $n \in \{1,2,3,4,5\}$ and cosine-similarity for obtaining the candidate scores.
    
    \item \textbf{ARC-I \cite{hu2014convolutional}:} This is a semantic matching model that uses CNN layers to compute the representation of the source and the candidate sequence, and then uses a feed-forward network to compute their similarity. 
    \item \textbf{ARC-II \cite{hu2014convolutional}:} This is an extension of ARC-I, which instead computes the interaction feature vector between the two sequences using CNN layers.
    \item \textbf{MV-LSTM \cite{wan2016deep}:} It is a neural semantic matching model that uses Bi-LSTM as encoder for both mention and candidate, and then computes an interaction vector using cosine similarity or a bilinear operation. 
    \item \textbf{MatchPyramid \cite{pang2016text}:} This model computes pair-wise dot product between mention and candidate to get an interaction matrix. The matrix is then passed through CNN layers with dynamic pooling to compute the similarity score.
    \item \textbf{KNRM \cite{xiong2017end}:} This is a neural ranking model which first computes the cosine similarity between each query word and document words. It then performs kernel pooling to compute the relevance score.
    \item \textbf{Duet \cite{mitra2017learning}:} It is a hybrid neural matching model that uses the word-level interaction and document-level similarity in a deep CNN architecture, to compute the similarity score.
    \item \textbf{Conv-KNRM \cite{dai2018convolutional}:} This model is an extension of KNRM, which instead uses convolutional layers to get $n$-gram representations of mention and candidates.
\end{itemize}

\begin{table}[!ht]
\centering
\begin{tabular}{p{2cm} p{1.15cm} p{1.15cm} p{1.15cm} p{1.15cm}}
\Xhline{3\arrayrulewidth}
 & \textbf{MedMentions} & & \textbf{3DNotes} & \\
\hline
\textbf{Model name} & \textbf{P@1} & \textbf{MAP} & \textbf{P@1} & \textbf{MAP} \\
\hline

TF-IDF & 61.39 & 67.74 & 56.89 & 69.45 \\
ARC-I  & 71.50 & 81.78 & 84.73 & 90.35 \\
ARC-II & 72.56 & 82.36 & 86.12 & 91.38 \\
KNRM   & 74.92 & 83.47 & 84.32 & 90.04 \\
Duet   & 76.19 & 84.92 & 86.11 & 91.19  \\
MatchPyramid & 78.15 & 86.31 & 85.97 & 91.32 \\
MV-LSTM      & 80.26 & 87.58 & 87.90 & 92.44 \\
Conv-KNRM    & 83.08 & 89.34 & 86.92 & 92.08 \\
\hline
LATTE-NKT    & 86.09 & 91.27 & 86.40 & 91.09 \\
\textbf{LATTE}    & \textbf{88.46} & \textbf{92.81} & \textbf{87.98} & \textbf{92.49} \\
\Xhline{3\arrayrulewidth}
\end{tabular}
\caption{Comparison of LATTE with other baseline models on MedMentions and 3DNotes dataset. LATTE-NKT is trained without the supervision of known types classification. P@1 is short for Precision@1.}
\label{tab:quantitative_results}
\end{table}

\subsection{Results}

\subsubsection{Quantitative Results}
Table \ref{tab:quantitative_results} shows the performance of LATTE against the state-of-the-art baselines. On MedMentions, LATTE outperforms the baselines by a wide margin. On 3DNotes, paired t-tests indicate that LATTE outperforms the strongest baseline with confidence level of 90\% (experimented with 5 different random seeds).

\noindent \textbf{Effect of using interaction-based method:} We can observe that TF-IDF and ARC-I, which compute the semantic representations of the mention and the candidate sequences independently, 
have lower performance as compared to all the other baselines. LATTE, as well as other models, use some form of interaction-based semantic representation of the two sequences. The interaction mechanism can model the pairwise relevance between the words from the two sequences, and hence, can uncover the relationship between them more effectively.

\noindent \textbf{Type modeling for entity linking:} We can see that all the other baselines, including LATTE-NKT, which is a version of our model without known type modeling, have lower performance than the full LATTE model. This shows that multi-tasking with type classification has strong positive effect on entity liking. Moreover, supervision on the known types guides the latent type modeling, which also contributes to the superior performance of LATTE. 

\noindent \textbf{Performance on datasets with different distributions:} MedMentions is from PubMed articles, which are more literary; 3DNotes is from dictated doctor's notes, which makes it colloquial in nature. The different distributions of the two datasets are also reflected in the out-of-vocabulary (OOV) words rate. Using GloVe embeddings, 3DNotes has 10.46\% OOV words, while MedMentions has 58.34\%. When the OOV rate is high, 1) character embeddings can help mitigate this problem as it captures the lexical similarity between words. 2) type information provides an extra link between a mention and the corresponding entity, which is beyond lexical and semantic matching. This explains why LATTE performs better on MedMentions than 3DNotes. Since typical biomedical datasets tend to have high OOV rate, we expect that the performance of LATTE on MedMentions can be generalized to that on other biomedical datasets.

\begin{figure*}[!ht]
\centering
\includegraphics[width=.95\textwidth]{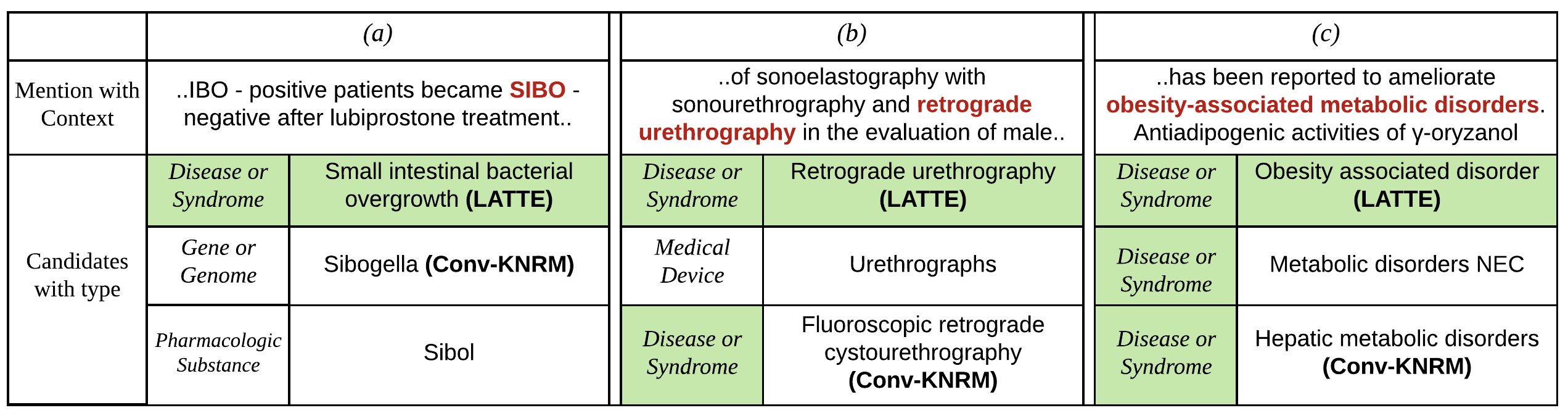}
\caption{Examples of entity linking result comparison between LATTE and a state-of-the-art baseline model (Conv-KNRM). Note that the red words are the mentions, and the green boxes are ground truth known types and candidates. (a) When candidates have different types, information of the correct mention type makes entity linking straightforward. LATTE learns how to classify mention types while doing entity linking. (b) When the mention or candidate words are out-of-vocabulary, measuring mention-candidate similarity becomes much harder. Character encoding and type classification mitigate this problem. (c) When candidates have the same type, LATTE is still capable of distinguishing the correct candidate from others.}
\label{fig:qualitative_results}
\end{figure*}

\begin{table}[!h]
\centering
\begin{tabular}{p{2.4cm} p{1.cm} p{1.cm} p{1.cm} p{1.cm}}
\Xhline{3\arrayrulewidth}
 & \textbf{MedMentions} & & \textbf{3DNotes} & \\
\hline
\textbf{Model name} & \textbf{P@1} & \textbf{MAP} & \textbf{P@1} & \textbf{MAP} \\
\hline
LATTE\_base & 80.02 & 86.94 & 84.08 & 90.15\\ 
LATTE\_base+LT & 86.09 & 91.27 & 86.40 & 91.09 \\
LATTE\_base+KT & 87.73 & 92.33 & 87.80 & \textbf{92.66} \\
LATTE & \textbf{88.46} & \textbf{92.81} & \textbf{87.98} & 92.49\\

\Xhline{3\arrayrulewidth}
\end{tabular}
\caption{Performance comparison of LATTE and its variants on MedMentions and 3DNotes Datasets.}
\label{tab:ablation}
\end{table}

\subsubsection{Ablation Analysis}
To study the effect of different components used in our model architecture, on the overall model performance, we also compare the performance of LATTE against its different variants (see Table \ref{tab:ablation}).

\begin{itemize}[leftmargin=*]
    \item \textbf{LATTE\_base:} This is the simplest variant of our model, which only contains the word embedding layer, encoder, the element-wise dot product as the similarity measure, and a feed-forward network to get a similarity score. 
    \item \textbf{LT:} This module includes the latent type encoder, softmax and the distribution similarity measure. From this step, character embedding is included and the mention-candidate interaction is switched to cross-attention. 
    \item \textbf{KT:} This module consists of the two known type classifiers, for mention and candidate respectively, depicted as the Known Type Classification layer in Figure \ref{fig:model_architecture}. Note that during test, we do not have the known type labels. 
\end{itemize}

As shown in Table \ref{tab:ablation}, 
introducing latent type modeling with cross-attention boosts Precision@1 on MedMentions and 3DNotes by 6.07\% and 2.32\% respectively, which shows that matching mention and candidates have similar latent type distribution, and modeling this similarity improves the entity linking task. It also shows that the cross-attention mechanism is strong in capturing the semantic similarity between mention and candidates. 
Instead, if we add the known type supervision, there are 7.71\% and 3.72\% gains in Precision@1 with respect to the two datasets. This shows that multi-tasking with known type classification has strong positive effect on the entity linking task. Finally, adding latent type modeling along with know type classification further improves the Precision@1. This proves that the hierarchical type modeling improves the entity linking task.  

\subsubsection{Qualitative Analysis}

Example in Figure \ref{fig:qualitative_results}(a) is a common case in biomedical domain, where the mention is an abbreviated form of the entity name. Such cases are challenging for traditional text matching methods since the abbreviation has very few common features with the complete name. Moreover, biomedical terms usually appear at a much lower frequency, and hence it is hard for models to learn the mapping through training. LATTE overcomes this problem by exploiting the type information. Although the mention may have a lower frequency, each type has a large amount of samples to train the type classifiers. Therefore our model can classify the mention type with higher confidence. If the candidates have different types, entity linking decision can be made with the knowledge of the type classification result. Note that, instead of direct usage, the type classification result is incorporated in the similarity computation. Figure \ref{fig:qualitative_results}(b) shows the case when the mention has OOV words. OOV words problem is a major challenge in the biomedical domain. Many biomedical terms do not have pre-trained word embeddings, without which the text matching becomes clueless. This is also why the retrieved result of the baseline model is incorrect. The character encoding and type matching in LATTE address this problem effectively. Figure \ref{fig:qualitative_results}(c) shows that when the candidates have the same type, LATTE can successfully distinguish the correct entity from other candidates. This is because: 1) the cross-attention mechanism is powerful in matching the mention and candidates text and 2) as discussed in previous sections, the latent types can be different even when the candidates share the same known type. Therefore the latent type modeling of LATTE works effectively in this case. 

\section{Conclusion}
\label{sec:conclusion}

We proposed a novel methodology, which models the latent type of mentions and entities, to improve the biomedical entity linking task. We incorporate this methodology in LATTE, a novel neural architecture that jointly performs fine-grained type learning and entity disambiguation. 
To the best of our knowledge, this is the first work to propose the idea of latent type modeling and apply it to biomedical entity linking. Our extensive set of experimental results shows that latent type modeling improves entity linking performance, and outperforms state-of-the-art baseline models. The idea of latent type modeling can be useful to a wider range, such as in other text matching tasks, and other non-biomedical domains. These can be possible directions for future work.

\section*{Acknowledgments}
\label{sec:ack}
This work was supported in part by the US National Science Foundation grants IIS-1619028 and IIS-1838730.

\bibliographystyle{aaai}
\bibliography{aaai-2020}

\end{document}